\documentclass[conference]{IEEEtran}
\IEEEoverridecommandlockouts
\usepackage{cite}
\usepackage{amsmath,amssymb,amsfonts}
\usepackage{algorithmic}
\usepackage{graphicx}
\usepackage{textcomp}
\usepackage{multirow}
\usepackage{subfigure}
\usepackage{xcolor}
\usepackage{amsmath}

\usepackage{booktabs}

\newcommand{\textFunction}[1]{\text{\textbf{\textsc{text}}}\left(#1\right)}

\def\BibTeX{{\rm B\kern-.05em{\sc i\kern-.025em b}\kern-.08em
    T\kern-.1667em\lower.7ex\hbox{E}\kern-.125emX}}
\begin{document}

\title{Forecasting Application Counts in Talent Acquisition Platforms: Harnessing Multimodal Signals using LMs
}

\author{\IEEEauthorblockN{Md Ahsanul Kabir}
\IEEEauthorblockA{
\textit{CareerBuilder LLC, USA}\\
mdahsanul.kabir@careerbuilder.com}
\and
\IEEEauthorblockN{Kareem Abdelfatah}
\IEEEauthorblockA{
\textit{CareerBuilder LLC, Canada}\\
kareem.abdelfatah@careerbuilder.com}
\and
\IEEEauthorblockN{Shushan He}
\IEEEauthorblockA{
\textit{CareerBuilder LLC, USA}\\
shushan.he@careerbuilder.com}
\and
\IEEEauthorblockN{Mohammed Korayem}
\IEEEauthorblockA{
\textit{CareerBuilder LLC, Canada}\\
mohammed.korayem@careerbuilder.com}
\and
\IEEEauthorblockN{ Mohammad Al Hasan}
\IEEEauthorblockA{
\textit{Indiana University Indianapolis, USA}\\
alhasan@iu.edu}
}

\maketitle

\begin{abstract}
As recruitment and talent acquisition have become more and more competitive, recruitment
firms have become more sophisticated in using machine learning (ML) methodologies for optimizing
their day to day activities. But, most of published ML based methodologies in this area have 
been limited to the tasks like candidate matching, job to skill matching, job classification 
and normalization. In this work, we discuss a novel task in the recruitment domain, namely,
application count forecasting, motivation of which comes from designing of effective outreach
activities to attract qualified applicants. We show that existing auto-regressive based time 
series forecasting methods perform poorly for this task. Henceforth, we propose a multimodal
LM-based model which fuses job-posting metadata of various modalities through a simple encoder. Experiments from large real-life datasets from CareerBuilder LLC show the effectiveness of the proposed method 
over existing state-of-the-art methods.
\end{abstract}

\section{Introduction}

Amidst the competitive talent acquisition landscape, a variety of research
efforts have focused on developing job recommendation
systems~\cite{cui2021workshop}, job-candidate matching~\cite{zhao2021embedding},
and creating shared latent spaces for job and skill
representation~\cite{al2024interactive}. However, relatively few studies address user
engagement with job postings—a metric strongly linked to recruiting agencies'
bottom line and valuable for crafting promotional strategies that attract
qualified candidates. To bridge this gap, CareerBuilder LLC is advancing
research on forecasting job application counts (JACs), a crucial measure of user
engagement. This paper introduces the task of JACs forecasting to the knowledge
discovery community, detailing our models and experimental results.

Forecasting JACs is central to recruitment analytics, offering insights to tailor recruitment strategies by predicting the number of applications a job may receive. This enables systems to respond proactively; for example, low application counts may trigger targeted marketing efforts or prompt a review of job descriptions. Predictive analytics also supports diversity and inclusion by highlighting demographic trends and providing tailored suggestions to diversify applicant pools.

Although forecasting is a well-established task in fields like retail and manufacturing, developing an accurate model for JACs forecasting is challenging. Job attributes comprise multimodal data—textual, categorical, and numerical—requiring robust representation learning to integrate these elements. Key challenges include effective entity extraction and schema standardization across typical job attributes like location~\cite{liu2019tripartite}, qualifications, skills, and salary. This complexity often leads to an extensive production pipeline. Auto-regressive time-series models, while an option, are limited as they use only historical counts, not external job features, as we demonstrate in our results.

Our approach to JACs forecasting is streamlined yet effective: we treat each data field (textual, categorical, or numerical) as a sentence, concatenate them into a paragraph, and input it into a pre-trained BERT model, leveraging BERT’s ability to process complex text structures. We show that this method outperforms traditional approaches requiring individual feature representations and complex feature fusion. By simplifying the pipeline and bypassing explicit entity extraction, this approach is efficient and highly manageable.

In summary, we claim three key contributions in this research work: First, we present a novel task, namely job applicant counts (JACs) forecasting. Second, we show that language models, like BERT can fuse multi-modal feature type seamlessly, thereby saves efforts of representation learning of features of different data types: numerical, categorical, or textual. 
Third, our extensive experiments demonstrate that our proposed job applicant counts method outperforms a large number of 
existing baselines by a significant margin.

\begin{figure}[t]
    \centering
    \includegraphics[scale=0.3]{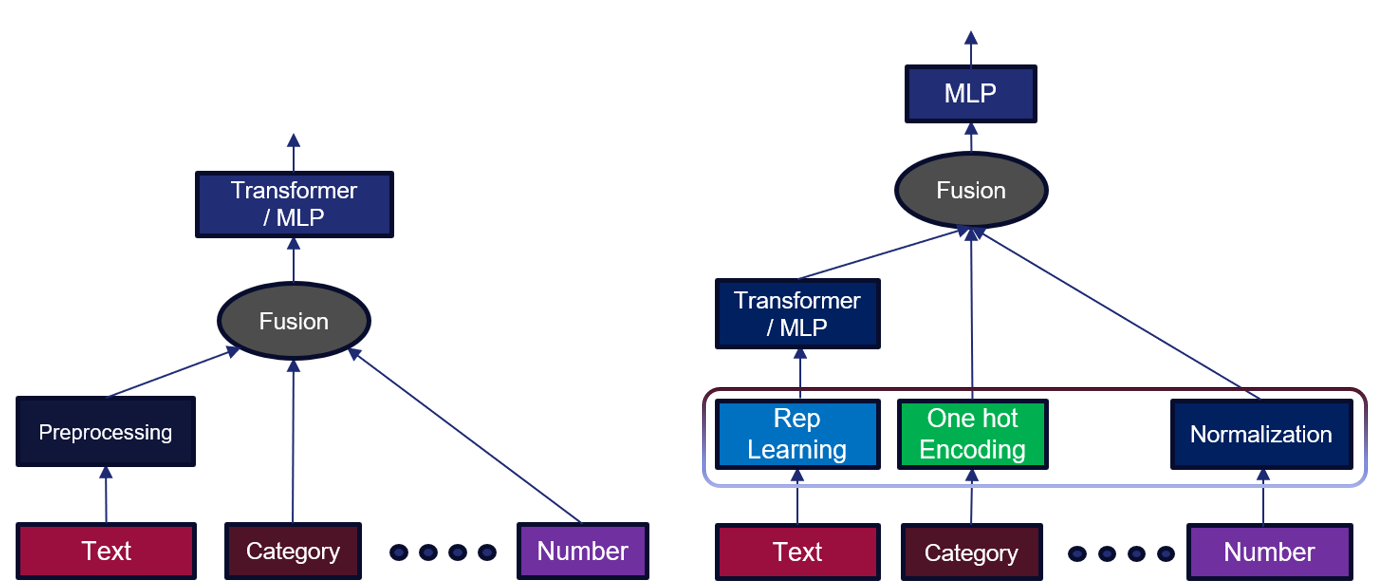}
    \caption{Feature Extraction Framework for the Experimental Methods; The left figure is the framework for LSTM with Spherical Embedding, Multimodal-BERT, GRU TSF and Multimodal-RoBERTa models while the right figure depicts the Multimodal-FF and Attention Fusion BERT models}
    \label{fig:classifier}
\end{figure}

\section{Problem Formulation}

Say, a dataset $\mathcal{D} = \{(x_i, y_i)\}_{i=1}^N$ is given, where $y_i$ is number of applicants for a job by time $t_i$ and $x_i$ comprises a blend of text ($\tau$), categorical ($c$), graph ($g$), location ($l$) and numerical ($n$) features associated with that job.
For instance, textual features, like job title, company name, and job description are included, alongside categorical features such as job type, state, channel, and job level. More richer features may also be available, such as: (1) the relationship between skills and jobs in the form of a bipartite graph; (2) job category presented as a job taxonomy tree, etc. Additionally, precise city locations can be translated into geolocation data. Each job has a duration of $t \in T$ days within the system along with the number of applicants that have applied for that job by that many days. 

By considering the task as a regression analysis, the JACs prediction task entails learning a function 
$f: (\mathbf{x}, T) \rightarrow \mathbb{R}$ that maps a vector from the job description space $\mathbf{x}$ and
time value $t \in T$ to a real number, an approximation of the number of applicants, $y$.  
The learning task is then to find $f^*$  that minimizes the function $\sum_{i=1}^N{\mathcal{L}(f^*((x_i, t_i), y_i))}$ over the dataset $\mathcal{D}$, where $\mathcal{L}$ is a suitable loss function quantifying the dissimilarity between the predicted and actual applicant counts.

\subsection{Methodology}
At CareerBuilder (CB), the existing pipeline predominantly focused on the conventional practice of feature extraction, treating each column separately, as illustrated in Figure~\ref{fig:feature-extraction-pipeline}.
CareerBuilder's current approach for application count forecasting task is built on top of this pipeline. 
Recently, we also developed a simple, yet effective method for application forecasting task on top of different language models, utilizing their superior textual comprehension ability. Our current approach is named as
Multimodal Feature Fusion (FF) and the recently developed method is called Multimodal Language Model (LM).

\subsubsection{Multimodal-FF} 
\begin{figure}[t]
    \centering
    \includegraphics[scale=0.32]{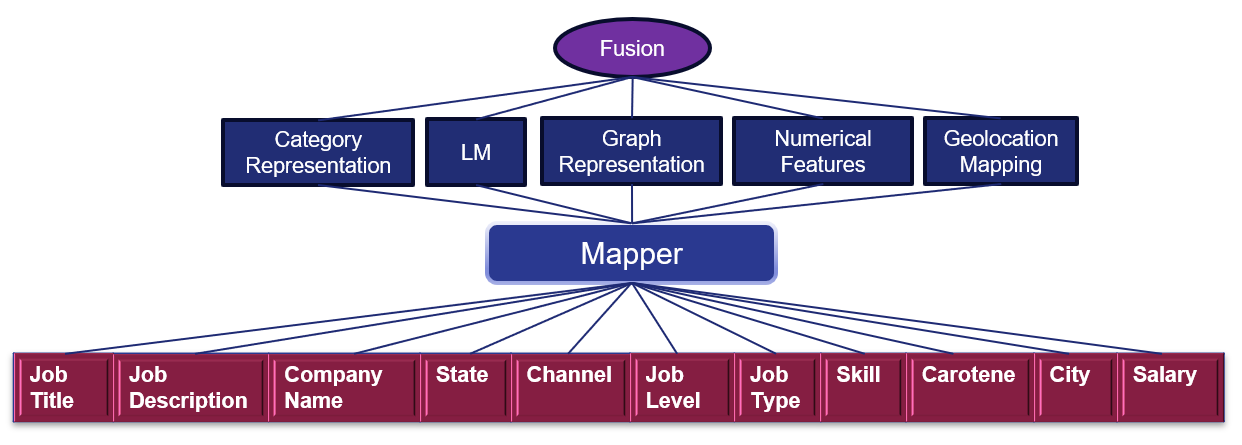}
    \caption{CareerBuilder's feature extraction pipeline: Each feature undergoes individual mapping by the mapper to apply its corresponding embedding approach, followed by the combination of all representations.}
    \label{fig:feature-extraction-pipeline}
\end{figure}

Multimodal-FF utilizes our proprietary embedding techniques to convert multimodal data of different types: categorical, graph, location, and textual to  high-dimensional real vector representation.
These vectors are then fused to create a representation for a given job, and the vector is fed into a multilayer perceptron for JAC forecasting. Figure~\ref{fig:feature-extraction-pipeline} illustrates the process for handling data from each modality. In this method, data from each modality is treated independently to represent it as a
vector. For example, when dealing with a location feature (city, state), the mapper identifies an appropriate embedding method for the location, and based on that calculates the associated location vector.

Let the embedding of the textual features is $E_t$, embedding of the categorical features is $E_c$, embedding of the graph nodes is $E_g$, and the embedding of the job location is $E_l$. $N$ is the remaining numerical feature values. The concatenated features, $\mathcal{F}_e$ is then expressed as: 

\[\mathcal{F}_e = E_t || E_c || E_g || E_l || N\]
The right side of the Figure ~\ref{fig:classifier} depicts the feature fusion state. The vectors coming from each modality is simply concatenated to form the embedding of a particular job. In the following paragraphs, we illustrate how we embed features in each modality.

For  text data, we employ a convolutional neural network (CNN)-based approach for text embedding~\cite{wang2019deepcarotene}, stemming from a multi-stream architecture. We leverage three text columns: company name, job title, and job description. Initially, we focus on the company name to construct a character-level embedding layer. Employing a stack of CNN layers, we extract features from the company name. Subsequently, a word-level embedding layer is applied to the job title, followed by feature extraction from each layer of a CNN stack. Similarly, features are extracted at the word level from job descriptions. The resulting concatenated vectors from each representation form the desired text embedding, $E_t$.

Within CareerBuilder, job and skill embedding involves the creation of a bipartite graph comprising jobs and skills, reflecting their direct associations. Additionally, two supplementary graphs, job transition, and skill transition, are established by observing user behavior, such as simultaneous application to two jobs or learning two skills. A method is then devised to embed jobs and skills, taking into account these interconnected graphs~\cite{Dave2018Represention}. The aim is to represent each job and skill so that similar and related skills or jobs demonstrate higher similarity scores compared to unrelated ones. This embedding process involves placing the jobs and skills into a $d$-dimensional latent space. To determine the correlation between a specific job and skill, all the skills associated with a job instance are embedded, and the mean pooling of all skill embeddings for that job is computed, resulting in the representation of the skill, denoted as $E_g$, for that particular job.

Job location is an important factor for candidates seeking a job, so the geographical location of a job holds significant importance for predicting application counts. However, traditional latitude and longitude coordinates pose a challenge due to their angular nature. To address this, we transform latitude and longitude into three-dimensional vectors using Cartesian coordinates, utilizing the equations~\cite{liu2019tripartite}: $x = cos \theta * cos \phi$, $y = cos \theta * sin \phi$, and $z = sin \theta$. Here, $\theta$ and $\phi$ denote the latitude and longitude of a location, respectively. This three-dimensional feature vector serves to embed the job's location, $E_l$.

Additionally, we include category features such as job type, state, channel, and job level. These features undergo one-hot encoding and are subsequently concatenated to form the category embedding, denoted as $E_c$. Furthermore, numerical features like salary, represented as $N$ are concatenated to form the final feature representation, $F_e$. Once the representation learning phase is completed, $F_e$ is stored as a database column for job instance embedding. To tackle the JAC forecasting task, we use a Multilayer Perceptron (MLP) by treating the forecasting as
a regression problem.

Embedding a job instance with various modalities separately is a common strategy, although it involves a complex procedure. As shown on the right side of Figure~\ref{fig:classifier}, the highlighted block signifies additional intricate tasks for feature extraction. Multiple models must be trained to acquire embeddings for each modality, and a mapper is required to discern the exact modality. Addressing unknown locations by relying on zero embedding introduces noise into the feature. Additionally, each modality is unaware of the embedding space of others; for instance, the text embedding method is not cognizant of job-skill representation, impacting the overall performance of a model. 

\subsubsection{Multimodal-LM} \label{sec:second} 

The prior feature extraction method is intricate, demanding entity extraction followed by the training of multiple models to address diverse modalities. To overcome this issue, we propose a different approach leveraging language models, which streamline the complex process by treating each modality as text-based sentence in a paragraph, often without employing an entity extraction. For example, numerical features like salary (or salary range) is generally available in job description; in this approach, we leave them as text in the job description paragraph expecting that a pretrained language model (LM) trained on vast corpora would automatically grasp the semantics of this information. This made job feature representation straightforward. As illustrated on the left side of Figure~\ref{fig:classifier}, the process involves the integration of converted text features into a paragraph, thereby eliminating the need for the explicit multimodal representation learning on different modality (the boxed part of right side of Figure~\ref{fig:classifier}). We call this process Multimodal-LM. 

\begin{table}
\caption{Distribution of the datasets for $t$=1, $t$=3, $t$=7, $t$=14, and $t$=30}
\centering
\small 
\setlength{\tabcolsep}{4pt}
\begin{tabular}{l | c | c | c }
\hline
\bf Day & \bf Train & \bf Test & \bf Val\\
\hline
Day 1 ($t$=1) & 109,163 & 20,321 &  19,003\\
Day 3 ($t$=3) & 267,678 & 59,603 &  47,703\\
Day 7 ($t$=7) & 305,440 & 62,169 &  53,930\\
Day 14 ($t$=14) & 239,117 & 55,971 & 42,152 \\
Day 30 ($t$=30) & 67,033 & 45,237 &  11,642\\
\hline
Combined & 988,431 & 243,301 &  174,430\\
\hline
\end{tabular}
\label{table:different-day-dataset}
\end{table}

Thus, for Multimodal-LM methods, the feature extraction technique involves converting all textual, categorical, graph node, location, and numerical features into textual representations. Categorical values simply becomes the text of category value instead of one-hot encoding,
numerical values becomes the ascii string of the digits in a number, etc. These representations are then
concatenated into a single unified text.
Formally, textual features $\tau$, categorical features ($c$), graph based features $g$, location features $l$, and numerical features ($n$) are all concatenated to $\mathcal{F}$ as below: 
\[\mathcal{F} = \tau + \textFunction{c} + \textFunction{g} +  \textFunction{l} + \textFunction{n}\]
where $\textFunction{c}$, and \textbf{+} denote casting a value into text and string-level concatenation operation respectively.
For the skills, we intentionally concatenate all the significant skills into a text sentence, which is the output of $\textFunction{g}$. 
When we combine all the features together, $\mathcal{F}$ becomes a text paragraph representing a job instance.

With one paragraph and a JAC value, we can readily rely on Language Models (LMs) to tackle the JAC forecasting task. We use both 'bert-base-uncased' pretrained model of BERT~\cite{Devlin2019BERTPO}, and the 'roberta-base' model of RoBERTa~\cite{liu2019roberta} to fine-tune for the JAC prediction regression task. We refer them as Multimodal-BERT and Multimodal-RoBERTa, respectively.

\section{Experiments and Results} 
We first discuss our dataset and the competing methods. Then we present the comparison between our proposed
methods and the competing methods. 

\subsection{Dataset Description}
CareerBuilder in-house dataset is used for this experiment. As we analyze the data, it is evident that 
job applicant count (JAC) follows a long-tail distribution. Majority of the jobs has 1-2 applicants,
and very few jobs have large applicant counts; all JAC values are less than 76. This count increases 
based on the number of days elapsed since the job has been posted, as can be seen in Table~\ref{table:different-day-dataset}. Shelf-life of many jobs are about 1 week, so in this
table $t=7$ days has the highest number of data instances. If we combine dataset of all days, there are approximately 1M instances in the train dataset, 174K in the validation dataset, and 243K in the test dataset, maintaining an approximate 8:2:2 ratio. 

\subsection{Description of the Competing Methods} 
\subsubsection{bi-LSTM and GRU-TSF} 
We employ bi-directional LSTM (bi-LSTM) and GRU Text Sequence Forecasting (TSF) approaches for JAC prediction. Both methods use Spherical Embedding for Text~\cite{meng2019spherical} with $d=100$ in the embedding layer and $L_1$ loss for regression.

\subsubsection{Attention Fusion BERT} 
Based on Gu and Budhkar~\cite{gu-budhkar-2021-package}, we implement Attention Fusion BERT, integrating multiple features through an attention mechanism~\cite{Bahdanau:2017:Attention}. We concatenate BERT text embeddings, one-hot encoded categories, and normalized numerical features, followed by a dense layer for regression.

\subsubsection{DeepTLF} 
DeepTLF~\cite{borisov2023deeptlf} addresses the challenge of applying DNNs to heterogeneous tabular data by using a TreeDrivenEncoder for feature transformation. It enhances predictive power by converting diverse inputs into homogeneous vectors. For this task, we encode job titles and descriptions using BERT and process categorical and numerical features with one-hot encoding and z-normalization.

\subsubsection{Time Series Forecasting} 
Job post time is crucial in predicting application trends, and time series models can be applied to forecast JACs. Using CareerBuilder's application history, we analyze time series data for jobs posted over at least 30 days. We test parametric models, such as exponential smoothing (SES), Croston, ADIDA, and IMAPA, using the statsforecast package~\cite{garza2022statsforecast} (results in Table~\ref{table:time-series-result}).

\subsection{Hyper-parameter Tuning} 
We utilize a single NVIDIA A100 GPU for all methods and implement them in PyTorch Python. Early Stopping with a tolerance of 5 is employed across all methods for hyperparameter tuning. Table ~\ref{table:hyperparameters} provides details on some hyperparameters for the competing method. LR, BS, RTPE illustrate Learning Rate, Batch Size and Run Time per Epoch respectively.

For the LSTM with Spherical Embedding method, the number of units in the LSTM block is tuned from 64 to 512 with an interval of 16, and the optimal performance is achieved when the number of units is set to 128. Similarly, the number of hidden units in the GRU-TSF is tuned, and the best results are obtained with 80 units. 

In the DeepTLF framework, BERT is employed for text embedding generation. The tunable parameters for DeepTLF include the number of hidden layers, estimators, hidden dimensions, batch size, xgb learning rate, and maximum depth. The number of hidden layers is adjusted to 64, 128, and 256, while the maximum depth is tuned from 3 to 4. Due to the memory demands of BERT embeddings, the batch size is set to 2. All other parameters are kept at their default settings as provided in the publicly available code.

We optimized the Multimodal-FF with four hidden layers consisting of 256, 128, 64, 32 neurons, respectively and varied the type of non-linear activation function in hidden layers, and the best non-linearity is given by ReLU activation based on validation set error. We initialize the weights with Glorot initialization \cite{Glorot2010UnderstandingTD} which protects gradients from explosion or vanishing.

Multimodal-BERT and Multimodal-RoBERTa employ a single dense unit for the Regression task, eliminating the need for MLP tuning. However, Attention Fusion BERT includes a single-layer MLP with variable units ranging from 64 to 512 (interval of 16) for the Regression task, with the best performance observed at 64 units. A dense layer follows this MLP for the Regression task.
Adam optimizer is used for LSTM with Spherical Embedding, GRU TSF, and Multimodal-FF with fixed learning rates of 1e-4, 1e-4, and 1e-3 respectively. For Multimodal-BERT, Multimodal-RoBERTa, and Attention Fusion BERT, AdamW optimizer is employed with learning rates of 1e-5 for all. Lastly, for DeepTLF, the learning rate refers to the XGBoost classifier's learning rate, set to the model's default value of 0.5.

\begin{table}
\caption{Hyper-parameters and Run Time of the Experimental Methods}
\begin{tabular}{l | c | c | c | c | c}
\hline
\bf Method & \bf LR &   \bf BS & \bf  Epochs &   \bf Inf-Time & \bf RTPE \\
\hline
LSTM & 1e-4  & 128 & 23 & 1.6 ms & 1.1 h\\
GRU TSF & 1e-4  & 128 & 25 & 1.2 ms & 0.75 h\\
DeepTLF & 0.5  & 2 & 25 & 2.3 ms &  1.5 h\\
Attention Fusion BERT & 1e-5  & 12 & 23 & 4.1 ms & 4.5 h\\
\midrule
Multimodal-FF &1e-3   & 256 & 50 & 1.1 ms&0.5 h \\
Multimodal-BERT & 1e-5  & 12 & 35 & 3.8 ms & 3.5 h\\
Multimodal-RoBERTa & 1e-5  & 12 & 20 & 3.9 ms & 4.2 h\\

\hline
\end{tabular}
\label{table:hyperparameters}
\end{table}

\begin{table*}
\caption{Performance (MAE / MALE) of Various Methods on Different Day Test Datasets Trained Separately}
\centering
\begin{tabular}{l | c | c | c | c | c}
\hline
\bf Method & \bf Day 1 & \bf Day 3 & \bf Day 7 & \bf Day 14 & \bf Day 30  \\
\hline
LSTM (Spherical Embedding) &  0.777 / 0.781 &  1.911 / 1.932 &  1.622 / 1.688 &  2.321 / 2.344 &  4.15 / 4.173 \\
DeepTLF &  0.787 / 0.789 &  2.102 / 1.992 &  1.7 / 1.712 &  2.222 / 2.347 &  3.922 / 3.999 \\
Attention Fusion BERT &  0.773 / 0.771 & 1.833 / 1.872 &  1.73 / 1.819 &  2.451 / 2.53 &  4.316 / 4.385 \\
\midrule
Multimodal-FF &  0.824 /0.822  &  \bf 1.146 /1.137  &  1.846 /1.841  &  2.217 /2.208  &  4.574 /4.562  \\
Multimodal-BERT &  \bf 0.498 / 0.497 &  1.683 / 1.621 & \bf 1.364 / 1.343 &  \ 1.58 / 1.571 &   3.914 / 3.869 \\

Multimodal-RoBERTa & 0.554 / 0.561& 1.711 / 1.713 & 1.357 / 1.352 & \bf 1.52 / 1.53 & \bf 3.87 / 3.91 \\
\hline
\end{tabular}
\label{table:separate-training}
\end{table*}

\begin{figure*}[t]
    \centering
    \begin{subfigure}{}
        \centering
        \includegraphics[scale=0.36]{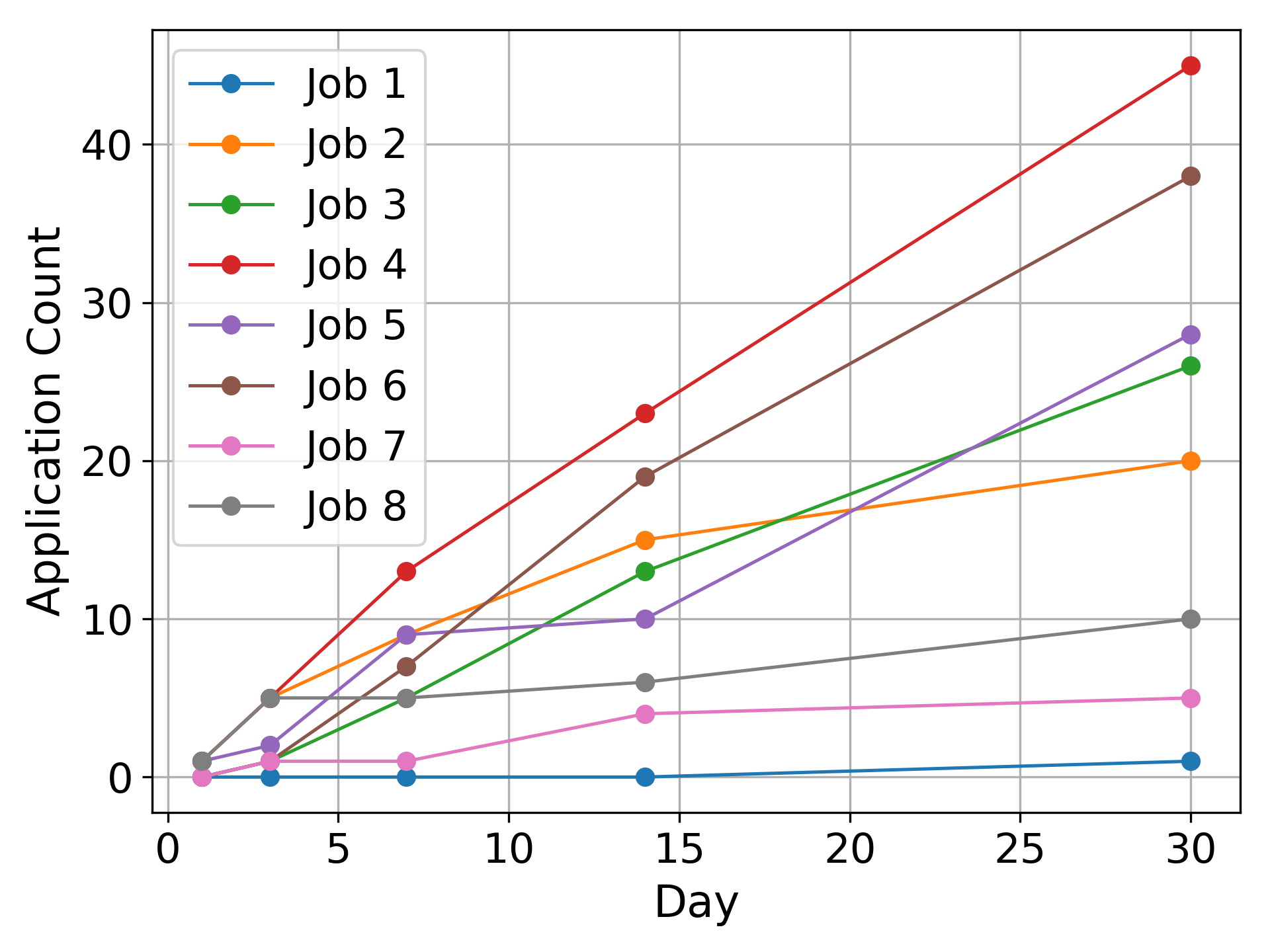}
        \label{subfig:left}
    \end{subfigure}
    \begin{subfigure}{}
        \centering
        \includegraphics[scale=0.36]{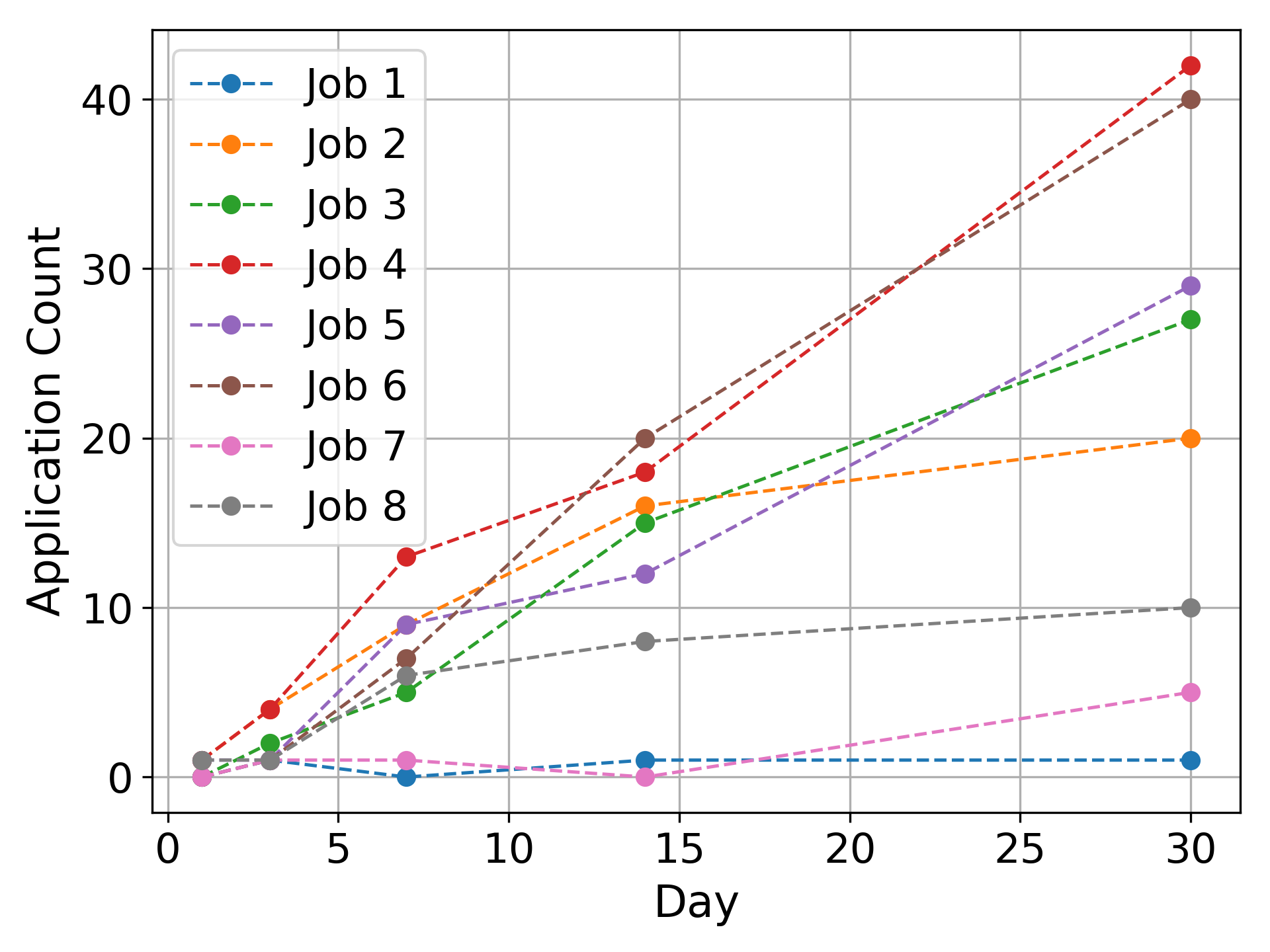}
        \label{subfig:right}
    \end{subfigure}
    \caption{Examples of some time series data samples in the dataset (left) and predictions using Multimodal-BERT (right)}
    \label{fig:ts-example}
\end{figure*}

\begin{table*}
\caption{Performance (MAE / MALE) of Various Methods on Different Day Test Datasets Trained Jointly}
\centering
\begin{tabular}{l | c | c | c | c | c | c}
\hline
\bf Method & \bf Day 1 & \bf Day 3 & \bf Day 7 & \bf Day 14 & \bf Day 30  & \bf overall \\
\hline
LSTM (Spherical Embedding) & \small 0.785 / 0.789 & \small  1.311 / 1.314 & \small 1.412 / 1.399 & \small  1.611/1.613 & \small 2.414/ 2.445 &\small  1.589 / 1.611 \\
DeepTLF  & \small 0.787 / 0.801 & \small  1.313 / 1.314 & \small 1.415 / 1.401 & \small  1.588/1.593 & \small 2.388/ 2.378 &\small  1.544 / 1.599 \\

GRU TSF & \small 0.766/ 0.768 & \small 1.297 / 1.299 & \small 1.422 / 1.402 & \small 1.589 / 1.612 & \small 2.244 / 2.235 & \small 1.535 / 1.438\\

Attention Fusion BERT & \small  0.811 / 0.812 & \small 1.627 / 1.631 & \small 1.683 / 1.713 & \small 2.521 / 2.533 & \small 4.115 / 4.185 & \small 2.489 / 2.394\\

\midrule

Multimodal-FF&  \small \bf 0.516 /\bf 0.513 &  \small 0.826 / 0.821 &  \small 1.444 / 1.439 &  \small 1.569 / 1.565 & \small  3.348 / 3.341 &  \small  1.598 / 1.593 \\
Multimodal-BERT & \small  0.521 / 0.514 & \small  \bf 0.811 / 0.805 & \small \bf 1.206 / 1.2 & \small \bf 1.099 / 1.093 & \small \bf 1.996 / 1.99 & \small  \bf 1.174 / 1.168 \\
Multimodal-RoBERTa & \small 0.554 / 0.561& \small 1.11 / 1.133 & \small  1.444 / 1.453 & \small  1.567 / 1.571 & \small  2.412 / 2.421 & \small  1.563/ 1.561\\
\hline
\end{tabular}
\label{table:joint-training}
\end{table*}

\begin{table*}
\caption{Performance (MAE / MALE) of Various Methods on Different JACs trained Jointly}
\centering
\begin{tabular}{l | c | c | c | c | c }
\hline
\bf Method & \bf JAC 1 & \bf JAC 3 & \bf JAC 5 & \bf JAC 7 & \bf JAC 9  \\
\hline
LSTM (Spherical Embedding) &  0.199 / 0.192 &  \bf 1.663 / 1.71 &  3.709 / 3.711 &  4.889 / 4.9 &  6.712 / 6.801 \\
DeepTLF &  0.189 / 0.19 &   1.667 / 1.683 &  3.714 / 3.713 &  4.901 / 4.9 &  6.677 / 6.703 \\
Attention Fusion BERT &  0.182 / 0.178 &  1.699 / 1.755 &  3.668 / 3.672 &  5.601 / 5.632 & 7.63 / 7.609 \\
\midrule
Multimodal-FF&  0.180/ 0.176 &   2.269 / 2.268 &  3.613 / 3.609 &  4.716 / 4.716 &  6.022 / 6.014 \\
Multimodal-BERT &  \bf 0.09/ 0.091 &  1.974 / 1.971 & \bf 3.141 / 3.134 & \bf 3.949 / 3.951 & \bf 4.78 / 4.793 \\

Multimodal-RoBERTa & 0.091 / 0.091  & 2.212 / 2.21 & 3.825 / 3.821 &  4.952 / 4.947 &  6.241 / 6.243 \\
\hline
\end{tabular}
\label{table:JAC-based-evaluation}
\end{table*}

\begin{table}
\caption{Time Series Forecasting Result for Day 30 Only}
\centering
\setlength{\tabcolsep}{3pt} 
\begin{tabular}{l | c | c}
\hline
\bfseries Method & \bfseries MAE & \bfseries MALE \\ 
\hline
ARIMA & 7.434 & 7.744 \\
AutoARIMA & 9.676 & 9.746 \\
AutoCES & 3.972 & 3.972 \\
AutoETS & 3.972 & 3.972 \\
AutoTheta & 24.507 & 24.170 \\
AutoRegressive(lags=[2]) & 3.972 & 3.972 \\
SES(alpha=0.5) & 6.070 & 6.470 \\
ADIDA & 7.628 & 7.905 \\
CrostonClassic & 8.854 & 9.180 \\
CrostonSBA & 8.916 & 9.255 \\
CrostonOptimized & 7.834 & 8.160 \\
TSB(0.3,0.2) & 8.094 & 8.415 \\
WindowAverage(window\_size=3) & 6.606 & 6.869 \\
\hline
\end{tabular}
\label{table:time-series-result}
\end{table}


Multimodal-FF stands out as the quickest method when considering inference time and run time per epoch. Its speed is credited to the utilization of pre-saved in-house representations and a straightforward MLP model, enabling efficient parallel computing. This distinguishes it from LSTM, DeepTLF, and GRU methods. Similarly, the separation of embedding and regression steps in Multimodal-BERT, Multimodal-RoBERTa, and Attention Fusion BERT also has the potential to diminish inference time, resembling the streamlined approach of Multimodal-FF. Although we use the BERT model to generate text embedding for DeepTLF, the BERT model itself is not trainable which makes it faster than multimodal-BERT and Multimodal-RoberTa.

As per hyper-parameter tuning with the time series models, we have fixed the hyperparameters as default or as illustrated in the Table~\ref{table:time-series-result}. In particular, automatic forecasting tools (eg. AutoARIMA, AutoCES, AutoTheta, AutoETS) search for the best parameters and possible non-seasonal model using Akaike Information Criterion (AICc). Considering the length of time series, short time observation window are tested for tuning. As results, 2 lags are included in the AutoRegression model and truncated series of size 3 are used for WindowAverage model. For the simple exponential smoothing (SES) model, the rate $alpha$ is set to 0.5. For Teunter-Syntetos-Babai (TSB) model has smoothing parameters 0.3, 0.2 for demand and probability, respectively.

\subsection{Results}
We compare all of our proposed methods Multimodal-FF, Multimodal-BERT, Multimodal-RoBERTa with the
competitors that we discussed above, namely LSTM with Spherical Embedding,  Attention Fusion BERT, DeepTLF, and GRU TSF and report the experimental results by using  Mean Average Error (MAE), and Mean Absolute Label Error
(MALE)~\cite{isir_ecom2022_8} metrics. In all results tables, the performance of our proposed methods is 
separated by a horizontal line.

Table~\ref{table:separate-training} showcases the performance of all methods trained individually on each day dataset. In this setup, each day has its own set of train, test, and validation data, as outlined in Table~\ref{table:separate-training}. As can be seen, among all the methods, our proposed Multimodal-FF exhibits superior performance for Day 3 data. Meanwhile, Multimodal-RoBERTa delivers the best performance for Day 14 and Day 30 datasets. However, for Day 1 and Day 7, Multimodal-BERT achieves the lowest MAE and MALE with values of 0.498, 0.497, and 1.364, 1.343, respectively, making it the top-performing method on these two days. None of the competing methods, like LSTM with Spherical Embedding or Attention Fusion BERT achieve
better results than any of our proposed methods for any of the days.

It is inconvenient to use a distinct model for predicting applicant counts for different days. To address this, we devised a consolidated dataset by integrating $t$ as an external feature, employing established techniques for feature extraction. While various methods exist for concatenating day features, our approach varies: for Attention Fusion BERT, DeepTLF, and Multimodal-FF, the day acts as an external feature encoded with one-hot encoding, while for other methods, including $t$, we adopt the Heterogeneous Feature Concatenation approach, incorporating all features in a single paragraph. Opting for a single model to streamline operations, we introduce GRU TSF for experimentation. Despite using a unified trained model, performance evaluation is conducted based on individual days. In the joint training results presented in Table~\ref{table:joint-training}, Multimodal-BERT outperforms other methods for all days except when $t = 1$, for which Multimodal-FF shows a slightly higher performance. However, as $t$ increases, Multimodal-BERT consistently outshines other baselines in terms of both MAE and MALE. The overall MALE of Multimodal-BERT is 23\% better than the second-best performing method, GRU TSF. The superior performance of Multimodal-BERT is surprising, as
this model does not perform any explicit representation learning for data of different modality, rather treat all data as textual feature in a paragraph. 

We also assess the performance of the jointly trained methods by grouping the instances based on the application count value. Within the test dataset, when JAC equals 1, Multimodal-BERT demonstrates the most superior performance. Surprisingly, for JAC equal to 3, the LSTM with Spherical Embedding model outperforms all other methods. Nevertheless, Multimodal-BERT consistently exhibits better performance than other methods across various JACs. One notable observation is that, as the data follows a long-tail distribution, predicting for larger JACs becomes more challenging. Despite this, the increase in JAC does not lead to a proportional increase in the Mean Absolute Label Error (MALE) of Multimodal-BERT compared to competing methods.

Given that the dataset we are analyzing exhibits characteristics of time-series data concerning the variable $t$, we conducted experiments utilizing various time-series methodologies as previously outlined. The left side of Figure~\ref{fig:ts-example} presents selected instances of the time series data from the test dataset, revealing a notable upward trend in JACs with the progression of $t$. The right side of Figure~\ref{fig:ts-example} shows the predictions of the example job instances using Multimodal-BERT. 

Conventional time series forecasting (TSF) models only consider past time stamps of the test dataset to predict for $t=30$. The overall performance of the time-series based methods is shown in Table~\ref{table:time-series-result}, with results exclusively provided for $t=30$ due to the necessity of observing a sufficient number of instances for effective prediction by time-series methods. Notably, CES, AutoETS, and the AutoRegressive methods collectively exhibit the best performance, all sharing the same MALE value of 3.972. It's essential to highlight that the overall performance of time-series methods are much poorer than the methods outlined in Table~\ref{table:joint-training}. This discrepancy could be attributed to the feature-agnostic nature of the time-series methods, which rely solely on the explored JACs of observed days. Additionally, the distinct day selection throughout the research paper results in a noticeable gap between $t=14$ and $t=30$. Depending solely on JACs for TSF does not seem promising based on our experimental findings.

\section{Conclusion}
In conclusion, in this paper, we introduce a novel task in the recruitment domain: job application count (JAC) forecasting. The significance of this work lies in its ability to address the challenges posed by the multifaceted nature of job postings, which contain textual, categorical, graph, location and numerical data. The proposed approach simplifies the feature extraction process and achieves superior performance compared to existing research works.

\bibliographystyle{IEEEtran}
\bibliography{CogSci_Template}

\begin{thebibliography}{10}
\providecommand{\url}[1]{#1}
\csname url@samestyle\endcsname
\providecommand{\newblock}{\relax}
\providecommand{\bibinfo}[2]{#2}
\providecommand{\BIBentrySTDinterwordspacing}{\spaceskip=0pt\relax}
\providecommand{\BIBentryALTinterwordstretchfactor}{4}
\providecommand{\BIBentryALTinterwordspacing}{\spaceskip=\fontdimen2\font plus
\BIBentryALTinterwordstretchfactor\fontdimen3\font minus \fontdimen4\font\relax}
\providecommand{\BIBforeignlanguage}[2]{{%
\expandafter\ifx\csname l@#1\endcsname\relax
\typeout{** WARNING: IEEEtran.bst: No hyphenation pattern has been}%
\typeout{** loaded for the language `#1'. Using the pattern for}%
\typeout{** the default language instead.}%
\else
\language=\csname l@#1\endcsname
\fi
#2}}
\providecommand{\BIBdecl}{\relax}
\BIBdecl

\bibitem{cui2021workshop}
X.~Cui, E.~Afshar, K.~Al-Jadda, S.~Kumar, J.~McAuley, T.~Ye, K.~Aryafar, V.~Dave, and M.~Korayem, ``Workshop on online and adaptative recommender systems (oars),'' in \emph{Proceedings of the 27th ACM SIGKDD Conference on Knowledge Discovery \& Data Mining}, 2021.

\bibitem{zhao2021embedding}
J.~Zhao, J.~Wang, M.~Sigdel, B.~Zhang, P.~Hoang, M.~Liu, and M.~Korayem, ``Embedding-based recommender system for job to candidate matching on scale,'' \emph{arXiv preprint arXiv:2107.00221}, 2021.

\bibitem{al2024interactive}
K.~Al~Jadda, M.~Korayem, B.~Tripp, A.~Soley, and S.~Proell, ``Interactive job recommendation and application submission tools of employment website entities,'' 2024.

\bibitem{liu2019tripartite}
M.~Liu, J.~Wang, K.~Abdelfatah, and M.~Korayem, ``Tripartite vector representations for better job recommendation,'' 2019.

\bibitem{wang2019deepcarotene}
J.~Wang, K.~Abdelfatah, M.~Korayem, and J.~Balaji, ``Deepcarotene-job title classification with multi-stream convolutional neural network,'' in \emph{2019 IEEE International Conference on Big Data (Big Data)}.

\bibitem{Dave2018Represention}
V.~S. Dave, B.~Zhang, M.~Al~Hasan, K.~AlJadda, and M.~Korayem, ``A combined representation learning approach for better job and skill recommendation,'' 2018.

\bibitem{Devlin2019BERTPO}
J.~Devlin, M.-W. Chang, K.~Lee, and K.~Toutanova, ``Bert: Pre-training of deep bidirectional transformers for language understanding,'' in \emph{NAACL}, 2019.

\bibitem{liu2019roberta}
Y.~Liu, M.~Ott, N.~Goyal, J.~Du, M.~Joshi, D.~Chen, O.~Levy, M.~Lewis, L.~Zettlemoyer, and V.~Stoyanov, ``Roberta: A robustly optimized bert pretraining approach,'' \emph{arXiv preprint arXiv:1907.11692}, 2019.

\bibitem{meng2019spherical}
Y.~Meng, J.~Huang, G.~Wang, C.~Zhang, H.~Zhuang, L.~Kaplan, and J.~Han, ``Spherical text embedding,'' \emph{Advances in neural information processing systems}, vol.~32, 2019.

\bibitem{gu-budhkar-2021-package}
K.~Gu and A.~Budhkar, ``A package for learning on tabular and text data with transformers,'' in \emph{Proceedings of the Third Workshop on Multimodal Artificial Intelligence}.\hskip 1em plus 0.5em minus 0.4em\relax Association for Computational Linguistics, 2021.

\bibitem{Bahdanau:2017:Attention}
D.~Bahdanau, K.~Cho, and Y.~Bengio, ``Neural machine translation by jointly learning to align and translate,'' 2015, 3rd International Conference on Learning Representations, ICLR 2015 ; Conference date: 07-05-2015 Through 09-05-2015.

\bibitem{borisov2023deeptlf}
V.~Borisov, K.~Broelemann, E.~Kasneci, and G.~Kasneci, ``Deeptlf: robust deep neural networks for heterogeneous tabular data,'' \emph{International Journal of Data Science and Analytics}, vol.~16, no.~1, pp. 85--100, 2023.

\bibitem{garza2022statsforecast}
C.~C. Federico~Garza, Max Mergenthaler~Canseco and K.~G. Olivares, ``Statsforecast: Lightning fast forecasting with statistical and econometric models,'' 2022.

\bibitem{Glorot2010UnderstandingTD}
X.~Glorot and Y.~Bengio, ``Understanding the difficulty of training deep feedforward neural networks,'' in \emph{International Conference on Artificial Intelligence and Statistics}, 2010.

\bibitem{isir_ecom2022_8}
M.~A. Kabir, M.~A. Hasan, A.~Mandal, D.~Tunkelang, and Z.~Wu, ``Ordsim: Ordinal regression for e-commerce query similarity prediction,'' in \emph{Proceedings of the International Workshop on Interactive and Scalable Information Retrieval methods for eCommerce (ISIR-eCom)}, 2022.

\end{thebibliography}
\end{document}